\documentclass[10pt,twocolumn,letterpaper]{article}

\usepackage[pagenumbers]{iccv} % To force page numbers, e.g. for an arXiv version

\usepackage{pifont}  
\usepackage{xcolor}
\newcommand{\cmark}{\ding{51}} 
 
\usepackage{algorithm}
\usepackage{algpseudocode}
\usepackage{multirow}

\definecolor{iccvblue}{rgb}{0.21,0.49,0.74}
\usepackage[pagebackref,breaklinks,colorlinks,allcolors=iccvblue]{hyperref}

\def\paperID{15} % *** Enter the Paper ID here
\def\confName{ICCV}
\def\confYear{2025}

\title{GLAD: Generalizable Tuning for Vision-Language Models}

\author{
Yuqi Peng$^{1,2}$
\qquad 
Pengfei Wang$^{3}$
\qquad
Jianzhuang Liu$^{1}$
\qquad
Shifeng Chen$^{1,4}$\thanks{Corresponding author: Shifeng Chen. This work was supported by the Shenzhen Science and Technology Program (JSGG20220831105002004) and the Shenzhen Key Laboratory of Computer Vision and Pattern Recognition.}
\vspace{0.5em}
\\
$^{1}$Shenzhen Institutes of Advanced Technology, Chinese Academy of Sciences \\
$^{2}$Northeastern University $^{3}$The Hong Kong Polytechnic University \\ $^{4}$Shenzhen University of Advanced Technology\\
{\tt\small peng.yuq@northeastern.edu, pengfei.wang@connect.polyu.hk,} \\ {\tt\small jz.liu@siat.ac.cn, shifeng.chen@siat.ac.cn  }
}
\begin{document}
\maketitle

\begin{figure*}
  \includegraphics[width=\textwidth]{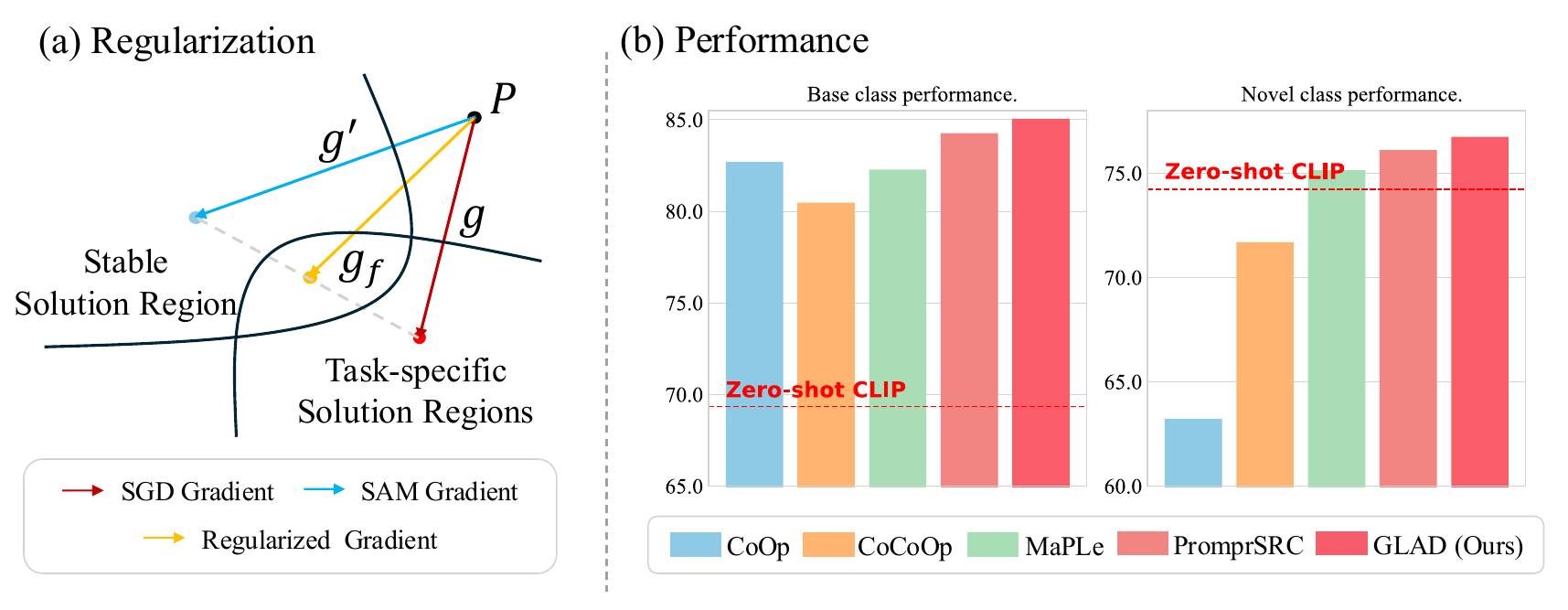}
  \vspace{-2em}
  \caption{\textnormal{An overview of the proposed regularization strategy and its effect on base-to-novel generalization. 
  (a) Illustration of our gradient regularization mechanism. We use the gradient computed via Sharpness-Aware Minimization (SAM) as a reference signal to steer the original gradient direction toward a region shared by both stable solutions and task-specific solutions. This regularization promotes convergence to parameter regions that are both robust and task-relevant. 
  (b) Comparison of GLAD with existing prompt learning methods in terms of base (seen during training) and novel (unseen during training) class accuracy across 11 datasets. GLAD consistently outperforms prior methods on both categories, demonstrating stronger generalization in few-shot settings.}}
  \label{fig:teaser}
  \vspace{-1.5em}
  
\end{figure*}

\begin{abstract}
Pre-trained vision-language models, such as CLIP, show impressive zero-shot recognition ability and can be easily transferred to specific downstream tasks via prompt tuning, even with limited training data. However, existing prompt tuning methods face two main challenges: (1) In few-shot scenarios, data scarcity often leads to overfitting, making the model sensitive to changes in the input domain. (2) To mitigate overfitting, these methods typically rely on complex task-specific model architectures and sensitive hyperparameter tuning, severely restricting their general applicability. To address these issues, we propose a simpler and more general framework called \textbf{GLAD} (\textbf{G}eneralizable \textbf{L}oRA tuning with Regul\textbf{A}rized Gra\textbf{D}ient). We show that merely applying LoRA achieves performance in downstream tasks comparable to current state-of-the-art prompt-based methods. While LoRA is effective and easy to use, it remains susceptible to overfitting in few-shot learning scenarios. To mitigate this risk, we introduce a gradient-based regularization technique. This technique effectively steers the optimization trajectory, encouraging the model to find a more stable parameter region that is robust to variations in data distribution. Through extensive experiments conducted on 15 benchmark datasets, we demonstrate that GLAD outperforms previous tuning approaches in terms of base-to-novel class generalization, image domain generalization, and cross-dataset generalization. The code will be publicly available.
\end{abstract}
   

\maketitle
\vspace{-0.5em}
\section{Introduction}
\vspace{-0.5em}
Foundational vision-language models have demonstrated remarkable capabilities across a variety of visual tasks, including object detection \cite{du2022learning,gu2022open,feng2022promptdet,zhong2022regionclip}, image classification \cite{radford2021learning,singh2022flava,zhai2022lit}, segmentation \cite{luddecke2022image}, and image description generation \cite{li2023blip,mokady2021clipcap,zhang2023llama}. Among these models, CLIP \cite{radford2021learning} has received significant attention due to its strong zero-shot recognition performance. CLIP employs contrastive learning on a large-scale dataset of image-text pairs sourced from the internet, mapping both visual and textual inputs into a shared embedding space. This approach endows CLIP with robust recognition capabilities in open-world settings, achieving strong performance even on rare data.

Although CLIP demonstrates outstanding performance across domains, adapting it to a specific downstream tasks under limited annotated data conditions remains a significant challenge \cite{zhou2022conditional}. Traditional full-parameter fine-tuning approaches often compromise the generalization capability acquired during pre-training, leading to catastrophic forgetting in few-shot setting. Prompt Tuning \cite{zhou2022learning, khattak2023maple,lee2023read, lu2022prompt, zhu2023prompt, zhou2022conditional, khattak2023self, yao2024tcp, zhang2024dept}, as a more parameter-efficient adaptation method, achieves few-shot adaptation by appending a set of learnable vectors to the input embeddings while keeping the backbone model frozen. However, this approach relies on few task-specific samples for optimizing the learnable vectors, making it prone to overfitting \cite{zhou2022learning, zhou2022conditional, lu2022prompt, zhang2024dept}. To address this, various methods \cite{lee2023read, khattak2023self} have proposed different prompt insertion\cite{zhou2022conditional} schemes, model structures\cite{lee2023read,khattak2023maple}, and objectives\cite{khattak2023self}. However, these often require task-specific architectures, complex regularization, or carefully tuned hyperparameters, limiting scalability to new tasks.

In this work, we propose GLAD, a simple yet effective fine-tuning framework designed to improve generalization in vision-language models. Unlike previous prompt tuning approaches \cite{zhou2022learning, khattak2023maple, lee2023read, lu2022prompt, zhu2023prompt, zhou2022conditional, khattak2023self, yao2024tcp}, GLAD leverages Low-Rank Adaptation (LoRA) \cite{hu2022lora}, which inserts trainable low-rank matrices into each transformer layer while keeping the pretrained parameters frozen. Surprisingly, we find that even a naive application of LoRA without any specific design can achieve performance comparable to recent state-of-the-art prompt tuning methods on downstream classification tasks.

Nevertheless, LoRA alone still has limitations in few-shot settings, where the scarcity of labeled data often leads to overfitting and makes the model less flexible in handling distributional shifts, resulting in poor generalization. To address this, we propose two complementary enhancements: a lightweight AlignNet module and a gradient-based regularization strategy. AlignNet dynamically adjusts textual features by conditioning on both visual and textual representations, enabling more adaptive text embeddings that better align with varying image input distributions. This improves the model’s flexibility in encoding semantics and strengthens its ability to generalize to new downstream task domains. In parallel, as illustrated in Figure~\ref{fig:teaser}(a), our gradient regularization technique steers the optimization direction toward stable regions in the parameter space, where the model’s performance is more robust to variations in data distribution. This leads to improved robustness under distribution shifts and helps prevent overfitting in few-shot settings. Figure~\ref{fig:teaser}(b) further illustrates that our method consistently outperforms prior approaches in base-to-novel generalization.

Our contributions are summarized as follows:
\begin{itemize}
\item We present \textbf{GLAD}, the first framework dedicated to improving the generalization ability of \textbf{LoRA} in few-shot learning scenarios. Our design enables efficient and scalable adaptation to new images, categories, and task domains under few-shot supervision.

\item We develop a novel \textbf{gradient-based regularization strategy} that steers the optimization direction toward flatter regions in the loss landscape. By aligning gradients with worst-case gradient directions, the method stabilizes optimization and mitigates overfitting under data scarcity, ultimately improving robustness to distribution shifts.

\item We conduct extensive classification experiments across three generalization settings: base-to-novel class generalization, image domain generalization, and cross-dataset generalization. GLAD consistently outperforms strong prompt tuning baselines and achieves state-of-the-art performance across all generalization tests, demonstrating its effectiveness in improving model performance and robustness under distributional shifts.
\end{itemize}
\vspace{-0.5em}

\section{Related Work}
\vspace{-0.5em}
In this section, we review three lines of related work that form the foundation of our method: prompt tuning in vision-language models, low-rank adaptation techniques for efficient fine-tuning, and regularization strategies for generalization.

\vspace{-0.5em}
\subsection{Prompt Tuning for Vision-Language Models} 
\vspace{-0.5em}
Prompt tuning has emerged as a parameter-efficient fine-tuning strategy, originally developed in natural language processing to guide pre-trained models using learnable prompt vectors \cite{lester2021power, liu2023pre}. In the context of VLMs, Zhou et al. \cite{zhou2022learning} proposed CoOp, which replaces handcrafted textual templates (e.g., ``a photo of a [CLASS]’’) with continuous learnable embeddings prepended to class tokens. While CoOp has achieved impressive performance in few-shot classification, its prompts tend to overfit the seen data distribution, leading to poor generalization ability.

To address overfitting in prompt tuning, many works introduce specialized designs. Some modify architectures to limit parameter updates and enhance flexibility, such as RPO \cite{lee2023read}, MaPLe \cite{khattak2023maple}, VPT \cite{jia2022visual}, and DePT \cite{zhang2024dept}. Others leverage prior knowledge from pre-trained models to guide learning, including PromptSRC \cite{khattak2023self}, PromptKD \cite{li2024promptkd}, TCP \cite{yao2024tcp}, and Quaternion Prompt \cite{cao2024domain}. Additionally, methods like CoCoOp \cite{zhou2022conditional} and MM-Align \cite{wang2023tuning} enhance cross-modal interactions to improve adaptation to distribution shifts.

In contrast, our approach eliminates the need for explicit prompt design by structurally modifying the model through LoRA. Additionally, we introduce a lightweight AlignNet that aligns the final text and visual features after encoding, enabling instance-level adaptability with minimal computational overhead.

\vspace{-0.5em}
\subsection{Low-Rank Adaptation for Vision-Language Models} 
\vspace{-0.5em}
LoRA, proposed by Hu et al. \cite{hu2022lora} as a low-rank parameter adaptation method for efficiently fine-tuning large language models. It injects trainable low-rank matrices into the weight layers of a Transformer while freezing the original weights, significantly reducing the number of tunable parameters and memory usage without sacrificing performance. This paradigm has since been extended to computer vision tasks \cite{liu2023visual, zhang2023adding, wang2024cogvlm, zhao2023svit, shen2024multimodal}, where LoRA and similar techniques have enabled efficient fine-tuning of both convolutional and Transformer-based models.

For vision-language models such as CLIP \cite{gao2021clip}, recent efforts have explored LoRA as an alternative to prompt tuning. CLIP-LoRA \cite{zanella2024low} demonstrates promising results in few-shot classification, outperforming several prompt-based approaches on task-specific benchmarks. However, their evaluation primarily focuses on tasks without domain shifts, leaving LoRA’s generalization ability largely unexplored. In contrast, our work is the first to systematically explore and evaluate the generalization potential of LoRA in vision-language fine-tuning, covering base-to-novel class generalization, image domain generalization, and cross-dataset generalization. This analysis not only complements existing studies but also provides a deeper understanding of LoRA’s strengths in few-shot VLM adaptation.

\vspace{-0.5em}
\subsection{Regularization Strategies} 
\vspace{-0.5em}
Regularization plays a critical role in improving model generalization, particularly in few-shot learning. Traditional regularization techniques can be broadly categorized into two families. The first includes constraint-based methods such as weight decay \cite{loshchilov2019decoupled,zhang2019three} and adversarial training \cite{goodfellow2014explaining}, which impose restrictions on model capacity or parameter updates to prevent overfitting. The second family comprises input- or parameter-dependent strategies, such as data augmentation \cite{zhang2017mixup,yun2019cutmix,verma2018manifold,uddin2021saliencymix,kim2021co,choi2023tokenmixup,park2021metropolis}, dropout \cite{srivastava2014dropout}, ensembling \cite{ilharco2022patching, wortsman2022robust}, and label smoothing \cite{szegedy2016rethinking, muller2019does}. These methods improve model robustness and performance in unseen scenarios by increasing diversity at the input level.

While effective, most existing techniques regularize either the input space or the optimization target. In contrast, our work introduces a gradient-based regularization method that directly steers the optimization trajectory. Specifically, we draw inspiration from Sharpness-Aware Minimization (SAM) \cite{foret2020sharpness}, which updates parameters based on the worst-case loss in the local neighborhood. Instead of directly adopting the perturbed gradient, we leverage the gradient computed by SAM as a correction signal to refine the original gradient. To maintain update stability, we remove components from the SAM gradient that conflict with the direction of the original gradient. This approach implicitly regularizes the training dynamics by guiding optimization toward a stable region in parameter space, corresponding to a flatter area in the loss landscape, which is empirically associated with better generalization \cite{foret2020sharpness, zhuang2022surrogate, wang2023sharpness, mordido2023lookbehind}. By integrating this regularization into the GLAD framework, we improve the model’s robustness against distribution shifts and overfitting, even under limited supervision.

\vspace{-0.5em}
\section{Method}
\vspace{-0.5em}
\begin{figure*}[t]
\centering
\includegraphics[width=\linewidth]{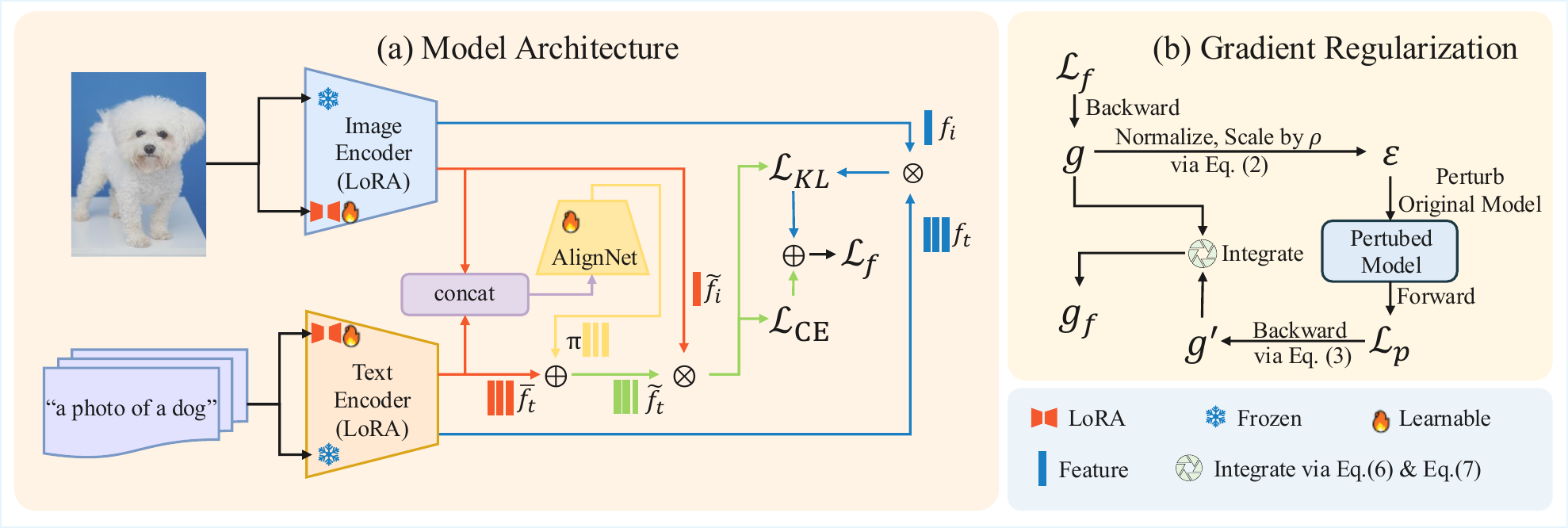}
\caption{\textnormal{Overview of the GLAD framework. (a) LoRA modules are inserted into each Transformer layer of CLIP’s encoders to enable efficient adaptation. A lightweight AlignNet adjusts text features using paired image features to enhance alignment. The blue features indicate outputs from the original CLIP model, the red features represent outputs after LoRA fine-tuning, and the green lines denote features adjusted by AlignNet. During training, we apply the cross-entropy loss $\mathcal{L}_{CE}$ for classification and the KL divergence loss $\mathcal{L}_{KL}$ between the predicted distributions of the original and fine-tuned models to preserve zero-shot CLIP behavior. (b) To improve generalization, a gradient regularization strategy fuses gradients from the original and perturbed models, guiding optimization toward a stable region.}}
\vspace{-0.5em}
\label{fig:overview}
\vspace{-0.5em}
\end{figure*}

We propose \textbf{GLAD} (Generalizable LoRA tuning with Regularized Gradient), a fine-tuning framework designed to address the limitations of existing fine-tuning methods in few-shot settings, where limited supervision often leads to overfitting and poor generalization to unseen domains. Our method consists of three main components: (1) \textit{LoRA-based internal adaptation baseline} (Sec.\ref{sec:lora}), which injects low-rank modules into each Transformer layer to enable parameter-efficient tuning; (2) \textit{visual-textual alignment via AlignNet} (Sec.\ref{sec:metanet}), which refines textual embeddings based on visual context to improve alignment; and (3) \textit{gradient-regularized optimization method} (Sec.\ref{sec:gradreg}), which stabilizes training by guiding optimization toward flatter regions in the loss landscape. Unlike previous prompt learning approaches that often rely on handcrafted model structures or complex regularization targets, GLAD enables direct representation adaptation with a simple yet effective design. An overview of the GLAD framework is illustrated in Figure~\ref{fig:overview}.
\vspace{-1.5em}
\subsection{Preliminaries}
\vspace{-0.5em}
Sharpness-Aware Minimization (SAM) \cite{foret2020sharpness} is a widely adopted optimization framework designed to improve the generalization capability of neural networks. Unlike standard empirical risk minimization, which seeks a parameter set $\boldsymbol{\theta}$ that minimizes the loss $\mathcal{L}(\boldsymbol{\theta})$ on training data, SAM explicitly promotes a stable region in the parameter space where small perturbations $\boldsymbol{\epsilon}$ to $\boldsymbol{\theta}$ result in minimal change in the loss. This corresponds to regions in the loss landscape that are relatively flat, and implies that the model is more robust to parameter perturbations and less sensitive to noise or input distributional shifts. By introducing adversarial perturbations to the model parameters during training and optimizing for performance under these perturbations, SAM encourages convergence to such stable regions. 

Formally, SAM formulates training as a min-max optimization problem:
\vspace{-0.5em}
\begin{equation}
\min_{\boldsymbol{\theta}} \ \max_{\| \boldsymbol{\epsilon} \|_2 \leq \rho} \ \mathcal{L}(\boldsymbol{\theta} + \boldsymbol{\epsilon}),
\label{eq:sam_obj}
\end{equation}

where $\boldsymbol{\epsilon}$ denotes an adversarial perturbation constrained within an $\ell_2$-norm ball of radius $\rho$, and $\mathcal{L}$ is the training loss function. This objective encourages the model to perform well not only at the current parameter point but also in its neighborhood, leading to solutions with improved robustness and better generalization.

To solve Eq.~\eqref{eq:sam_obj} efficiently, SAM approximates the inner maximization via a first-order expansion and computes the perturbation direction as:
\vspace{-0.5em}
\begin{equation}
\boldsymbol{\epsilon} = \rho \cdot \frac{\nabla_{\boldsymbol{\theta}} \mathcal{L}(\boldsymbol{\theta})}{\| \nabla_{\boldsymbol{\theta}} \mathcal{L}(\boldsymbol{\theta}) \|_2},
\label{eq:sam_perturb}
\end{equation}
followed by computing the updated gradient at the perturbed parameters:
\begin{equation}
\nabla_{\text{SAM}} \mathcal{L}(\boldsymbol{\theta}) = \nabla_{\boldsymbol{\theta}} \mathcal{L}(\boldsymbol{\theta} + \boldsymbol{\epsilon}).
\label{eq:sam_grad}
\end{equation}
\vspace{-0.5em}
The resulting gradient reflects the worst-case local behavior.

SAM has demonstrated remarkable performance in standard supervised learning. Its ability to improve robustness against perturbations motivates our use of gradient-based regularization in few-shot learning scenarios. In Section~\ref{sec:gradreg}, we present a gradient-based regularization method that uses SAM gradients to guide the optimization direction, improving performance in few-shot settings while enhancing the model’s stability to input distribution shifts and its overall generalization ability.

\vspace{-0.2em}
\subsection{Internal Representation Tuning via LoRA}
\vspace{-0.2em}
\label{sec:lora}
GLAD leverages Low-Rank Adaptation (LoRA) modules to enable efficient tuning of internal representations in both the image and text encoders of CLIP, targeting key linear projections such as the query, key, and value matrices in self-attention. This hierarchical integration enables layer-wise adaptation of intermediate features, facilitating a deeper and more distributed shift in representations toward downstream tasks.

For an input $\boldsymbol{x}$, a hidden state $\boldsymbol{h}$, and a weight matrix $\mathbf{W} \in \mathbb{R}^{M \times N}$, each LoRA module introduces a low-rank residual $\Delta \mathbf{W} = \mathbf{B} \mathbf{A}$, where $\mathbf{A} \in \mathbb{R}^{r \times N}$ and $\mathbf{B} \in \mathbb{R}^{M \times r}$ with rank $r \ll \min(M, N)$. The output is computed as:
\begin{equation}
\boldsymbol{h} = \mathbf{W} \boldsymbol{x} + \gamma \Delta \mathbf{W} \boldsymbol{x} = \mathbf{W} \boldsymbol{x} +\gamma  \mathbf{B} (\mathbf{A} \boldsymbol{x}),
\end{equation}
where $\gamma$ is a scaling factor, $\mathbf{W}$ is frozen during training, and only the low-rank matrices $\mathbf{A}$ and $\mathbf{B}$ are updated. 

This enables GLAD to efficiently adapt to new tasks by introducing only about $1\%$ additional parameters, without forgetting the knowledge stored in the pretrained weights. As shown in Table~\ref{tab:ablations}, this design preserves the advantages of parameter efficiency while significantly improving downstream task adaptability and maintaining robustness to input distribution shifts. An illustration of this architecture is shown in Figure~\ref{fig:overview}(a). Additionally, unlike prompt learning, which introduces extra input tokens, the LoRA modules can be merged into the original weight matrices ($\mathbf{W}_{\mathrm{merged}} = \mathbf{W} + \gamma \Delta \mathbf{W}$) during inference, improving inference efficiency.

\vspace{-0.2em}
\subsection{Image-Aware Text Feature Adjustment via AlignNet}
\vspace{-0.2em}
\label{sec:metanet}
In the dual-encoder architecture of CLIP, images and texts are encoded independently through separate encoders. While this design allows for efficient inference—since text embeddings can be precomputed and reused—it also limits the model’s flexibility in adapting to specific downstream tasks, where static textual representations may fail to align accurately with dynamic visual features.

To enhance task adaptability while maintaining computational efficiency, we propose a novel AlignNet module that performs post-encoding text adjustment based on static textual and dynamic visual features. The workflow of AlignNet is illustrated in Figure~\ref{fig:overview}(a). Specifically, given a dynamic image feature $\tilde{\mathbf{f}_i} \in \mathbb{R}^{512}$ and a static text feature $\bar{\mathbf{f}_t} \in \mathbb{R}^{512}$, we concatenate them to form a joint representation $[\tilde{\mathbf{f}_i}; \bar{\mathbf{f}_t]} \in \mathbb{R}^{1024}$, which is passed through a lightweight AlignNet (MLP) to generate a bias vector $\boldsymbol{\pi} \in \mathbb{R}^{512}$. The adjusted text embedding is computed as:
\vspace{-0.5em}
\begin{equation}
\tilde{\mathbf{f}}_t
= \bar{\mathbf{f}_t} + \boldsymbol{\pi}
= \bar{\mathbf{f}_t} + \text{AlignNet}([\tilde{\mathbf{f}_i}; \bar{\mathbf{f}_t}]).
\end{equation}
\vspace{-1.5em}

By directly operating on the final text embeddings, AlignNet not only maintains computational efficiency, since textual features do not require re-encoding at inference time, but also enhances the model’s ability to align visual and textual semantics in a more flexible and expressive manner. The empirical results in Table~\ref{tab:ablations} demonstrate that this adjustment significantly improves performance in downstream tasks without compromising generalizability.

\begin{algorithm}[t]
\caption{Gradient Regularization Algorithm}
\label{alg:grad_reg}
\begin{algorithmic}[1]
\Require Model parameters $\boldsymbol{\theta}$, loss $\mathcal{L}$, perturbation radius $\rho$, mixing factor $\alpha \in [0,1]$, small constant $\delta$
\Ensure Regularized gradient $\mathbf{g}_f$
\vspace{2pt}

\State Compute original gradient: $\mathbf{g} \gets \nabla_{\boldsymbol{\theta}} \mathcal{L}(\boldsymbol{\theta})$
\State Compute perturbation $\boldsymbol{\epsilon}$ using Eq.~(\ref{eq:sam_perturb})
\State Compute perturbed gradient: $\mathbf{g}' \gets \nabla_{\boldsymbol{\theta}} \mathcal{L}(\boldsymbol{\theta} + \boldsymbol{\epsilon})$ \hfill \textit{(Eq.~\ref{eq:sam_grad})}
\vspace{2pt}

% \State $s \gets \langle \mathbf{g}, \mathbf{g}' \rangle$
\If{$\langle \mathbf{g}, \mathbf{g}' \rangle < 0$}
    \State $\mathbf{g}' \gets$ projection of $\mathbf{g}'$ onto subspace orthogonal to $\mathbf{g}$ \hfill \textit{(Eq.~\ref{eq:proj_filter})}
\EndIf

\State Interpolate: $\mathbf{g}_f \gets (1 - \alpha)\mathbf{g} + \alpha \mathbf{g}'$ \hfill \textit{(Eq.~\ref{eq:final_grad})}

\State \Return $\mathbf{g}_f$
\end{algorithmic}
\end{algorithm}
\vspace{-0.5em}
\subsection{Gradient-Regularized Optimization for Flat Minima}
\vspace{-0.2em}
\label{sec:gradreg}

Few-shot tuning is highly prone to overfitting due to the limited number of training samples. In such settings, models tend to memorize the training data and exhibit high sensitivity to input variations, leading to poor generalization on unseen distributions. To alleviate this issue, we propose a gradient-regularized optimization strategy that encourages convergence toward flatter and more stable regions of the loss surface.

Recent studies have highlighted that models with sharp minima tend to exhibit degraded performance when exposed to input or domain shifts \cite{foret2020sharpness, zhuang2022surrogate, wang2023sharpness, mordido2023lookbehind}. In contrast, solutions that lie in flatter regions of the loss landscape are empirically associated with better generalization.

Since Sharpness-Aware Minimization (SAM) has demonstrated strong performance in finding flatter loss landscapes, we first explore its effectiveness in few-shot learning scenarios. As shown in Table~\ref{tab:ablations}, we find that while SAM helps preserve CLIP’s original generalization ability, it struggles to adapt to downstream tasks using only a small number of samples. To address this, we design a regularization strategy that leverages the perturbed gradient direction as a reference without altering the main optimization direction. Our goal is to guide the model toward flatter regions of the loss landscape while preserving the descent direction of the original gradient. The workflow of this regularization process is illustrated in Figure~\ref{fig:overview}(b).

Specifically, let $\mathbf{g}$ denote the gradient of the loss with respect to model parameters, and let $\mathbf{g}’$ be a reference gradient. We first compute the inner product $\langle \mathbf{g}, \mathbf{g}’ \rangle$. If the two gradients exhibit conflicting directions (i.e., $\langle \mathbf{g}, \mathbf{g}’ \rangle < 0$), we remove the opposing component from $\mathbf{g}’$ by projecting it onto the subspace orthogonal to $\mathbf{g}$:
\vspace{-0.2em}
\begin{equation}
\mathbf{g}' \leftarrow \mathbf{g}' - \frac{\langle \mathbf{g}, \mathbf{g}' \rangle}{\|\mathbf{g}\|^2 + \delta} \cdot \mathbf{g}, \quad \text{if } \langle \mathbf{g}, \mathbf{g}' \rangle < 0,
\label{eq:proj_filter}
\end{equation}
\vspace{-0.2em}
where $\delta$ is a small constant to ensure numerical stability.

We then compute the final update direction by linearly interpolating between the original and adjusted gradients:
\vspace{-0.2em}
\begin{equation}
\mathbf{g}_f = (1 - \alpha)\mathbf{g} + \alpha \mathbf{g}',
\label{eq:final_grad}
\end{equation}
\vspace{-0.2em}
where $\alpha \in [0, 1]$ is a mixing coefficient.

This fused gradient $\mathbf{g}_f$ combines the original gradient with a smoothed reference gradient, guiding the optimization toward flatter regions of the loss landscape. In practice, this improves robustness to input distribution shifts while preserving learning capability in few-shot scenarios. The full procedure is outlined in Algorithm~\ref{alg:grad_reg}.

\vspace{-0.2em}
\section{Experiments}
\vspace{-0.2em}
We conduct extensive experiments to evaluate the effectiveness and generalization ability of our proposed GLAD under three settings: base-to-novel class generalization, image domain generalization, and cross-dataset generalization, following \cite{huang2022learning, zhou2022conditional}. In the subsequent sections, we present quantitative results and ablation studies to analyze the contributions of each component in GLAD.

\begin{table*}[t!]
\centering
\addtolength{\tabcolsep}{+6pt}
\resizebox{\textwidth}{!}{%
\begin{tabular}{lc|c c c c c c c c | c c}
\toprule
\multirow{2}{*}{\centering Dataset} &  &  {CLIP} &  {CoOp} & {CoCoOp} &  {ProDA} & {RPO} & {MaPLe} & {PromptSRC} & {CLIP-LoRA} & {GLAD} & Gain \\
 & &  \cite{radford2021learning} & \cite{zhou2022learning} & \cite{zhou2022conditional}  & \cite{lu2022prompt} & \cite{lee2023read} & \cite{khattak2023maple}  & \cite{khattak2023self} & \cite{zanella2024low} & (Ours) & ($\Delta$) \\
\midrule
\multirow{3}{*}{\shortstack[l]{Average on \\ 11 datasets}} 
    & Base & 69.34 & 82.69 & 80.47 & 81.56 & 81.13 & 82.28 & 84.26 & \underline{84.47} & \textbf{85.05} & \textcolor{blue}{+0.58} \\
    & Novel & 74.22 & 63.22 & 71.69 & 72.30 & 75.00 & 75.14 & \underline{76.10} & 74.22 & \textbf{76.74} & \textcolor{blue}{+2.52} \\
    & HM & 71.70 & 71.66 & 75.83 & 76.65 & 77.94 & 78.55 & \underline{79.97} & 79.01 & \textbf{80.68} & \textcolor{blue}{+1.67} \\
\midrule
\multirow{3}{*}{ImageNet} 
    & Base & 72.43 & 76.47 & 75.98 & 75.40 & 76.60 & 76.66 & 77.60 & \underline{77.73} & \textbf{79.03} & \textcolor{blue}{+1.30} \\
    & Novel & 68.14 & 67.88 & 70.43 & 70.23 & \textbf{71.57} & 70.54 & 70.73 & 69.17 & \underline{71.43} & \textcolor{blue}{+2.26} \\
    & HM & 70.22 & 71.92 & 73.10 & 72.72 & 74.00 & 73.47 & \underline{74.01} & 73.20 & \textbf{75.04} & \textcolor{blue}{+1.84} \\
\midrule
\multirow{3}{*}{Caltech101} 
    & Base & 96.84 & 98.00 & 97.96 & \textbf{98.27} & 97.97 & 97.74 & \underline{98.10} & 97.83 & 98.20 & \textcolor{blue}{+0.37} \\
    & Novel & 94.00 & 89.81 & 93.81 & 93.23 & \underline{94.37} & 94.36 & 94.03 & 93.83 & \textbf{94.83} & \textcolor{blue}{+1.00} \\
    & HM & 95.40 & 93.73 & 95.84 & 95.68 & \underline{96.14} & 96.02 & 96.02 & 95.79 & \textbf{96.49} & \textcolor{blue}{+0.70} \\
\midrule
\multirow{3}{*}{OxfordPets} 
    & Base & 91.17 & 93.67 & 95.20 & \underline{95.43} & 94.63 & \underline{95.43} & 95.33 & 95.33 & \textbf{95.87} & \textcolor{blue}{+0.54} \\
    & Novel & 97.26 & 95.29 & 97.69 & \textbf{97.83} & 97.50 & 97.76 & 97.30 & \underline{97.77} & 97.70 & \textcolor{red}{-0.07} \\
    & HM & 94.12 & 94.47 & 96.43 & 96.62 & 96.04 & \underline{96.58} & 96.30 & 95.53 & \textbf{96.78} & \textcolor{blue}{+1.25} \\
\midrule
\multirow{3}{*}{Stanford Cars} 
    & Base & 63.37 & 78.12 & 70.49 & 74.70 & 73.87 & 72.94 & \underline{78.27} & 81.23 & \textbf{81.60} & \textcolor{blue}{+0.37} \\
    & Novel & 74.89 & 60.40 & 73.59 & 71.20 & \textbf{75.53} & 74.00 & \underline{74.97} & 71.93 & 74.93 & \textcolor{blue}{+3.00} \\
    & HM & 68.65 & 68.13 & 72.01 & 72.91 & 74.69 & 73.47 & \underline{76.58} & 76.30 & \textbf{78.12} & \textcolor{blue}{+1.82} \\
\midrule
\multirow{3}{*}{Flowers102} 
    & Base & 72.08 & 97.60 & 94.87 & 97.70 & 94.13 & 95.92 & \underline{98.07} & 97.70 & \textbf{98.33} & \textcolor{blue}{+0.63} \\
    & Novel & \textbf{77.80} & 59.67 & 71.75 & 68.68 & \underline{76.67} & 72.46 & 76.50 & 72.63 & 75.40 & \textcolor{blue}{+2.77} \\
    & HM & 74.83 & 74.06 & 81.71 & 80.66 & 84.51 & 82.56 & \textbf{85.95} & 83.32 & \underline{85.35} & \textcolor{blue}{+2.03} \\
\midrule
\multirow{3}{*}{Food101} 
    & Base & 90.10 & 88.33 & \underline{90.70} & 90.30 & 90.33 & 90.71 & 90.67 & 90.07 & \textbf{90.90} & \textcolor{blue}{+0.83} \\
    & Novel & 91.22 & 82.26 & 91.29 & 88.57 & 90.83 & \textbf{92.05} & 91.53 & 91.00 & \underline{92.03} & \textcolor{blue}{+1.03} \\
    & HM & 90.66 & 85.19 & 90.99 & 89.43 & 90.58 & \underline{91.38} & 91.10 & 90.53 & \textbf{91.46} & \textcolor{blue}{+0.93} \\
\midrule
\multirow{3}{*}{FGVC Aircraft} 
    & Base & 27.19 & 40.44 & 33.41 & 36.90 & 37.33 & 37.44 & \underline{42.73} & 43.80 & \textbf{43.90} & \textcolor{blue}{+0.10} \\
    & Novel & 36.29 & 22.30 & 23.71 & 34.13 & 34.20 & 35.61 & \underline{37.87} & 34.47 & \textbf{38.33} & \textcolor{blue}{+3.86} \\
    & HM & 31.09 & 28.75 & 27.74 & 35.46 & 35.70 & 36.50 & \underline{40.15} & 38.58 & \textbf{40.93} & \textcolor{blue}{+2.35} \\
\midrule
\multirow{3}{*}{SUN397} 
    & Base & 69.36 & 80.60 & 79.74 & 78.67 & 80.60 & 80.82 & \underline{82.67} & 82.03 & \textbf{83.23} & \textcolor{blue}{+1.20} \\
    & Novel & 75.35 & 65.89 & 76.86 & 76.93 & 77.80 & \underline{78.70} & 78.47 & 77.67 & \textbf{79.77} & \textcolor{blue}{+2.10} \\
    & HM & 72.23 & 72.51 & 78.27 & 77.79 & 79.18 & 79.75 & \underline{80.52} & 79.79 & \textbf{81.46} & \textcolor{blue}{+1.67} \\
\midrule
\multirow{3}{*}{DTD} 
    & Base & 53.24 & 79.44 & 77.01 & 80.67 & 76.70 & 80.36 & \underline{83.37} & 83.00 & \textbf{84.43} & \textcolor{blue}{+1.43} \\
    & Novel & 59.90 & 41.18 & 56.00 & 56.48 & 62.13 & 59.18 & \underline{62.97} & 59.33 & \textbf{64.03} & \textcolor{blue}{+4.70} \\
    & HM & 56.37 & 54.24 & 64.85 & 66.44 & 68.65 & 68.16 & \underline{71.75} & 69.20 & \textbf{72.83} & \textcolor{blue}{+3.63} \\
\midrule
\multirow{3}{*}{EuroSAT} 
    & Base & 56.48 & 92.19 & 87.49 & 83.90 & 86.63 & \underline{94.07} & 92.90 & \textbf{94.30} & 91.83 & \textcolor{red}{-2.47} \\
    & Novel & 64.05 & 54.74 & 60.04 & 66.00 & 68.97 & 73.23 & \underline{73.90} & 70.63 & \textbf{77.77} & \textcolor{blue}{+7.14} \\
    & HM & 60.03 & 68.69 & 71.21 & 73.88 & 76.80 & \underline{82.35} & 82.32 & 80.77 & \textbf{84.22} & \textcolor{blue}{+3.45} \\
\midrule
\multirow{3}{*}{UCF101} 
    & Base & 70.53 & 84.69 & 82.33 & 85.23 & 83.67 & 83.00 & \underline{87.10} & 86.20 & \textbf{88.13} & \textcolor{blue}{+1.93} \\
    & Novel & 77.50 & 56.05 & 73.45 & 71.97 & 75.43 & 78.66 & \underline{78.80} & 78.03 & \textbf{80.03} & \textcolor{blue}{+2.00} \\
    & HM & 73.85 & 67.46 & 77.64 & 78.04 & 79.34 & 80.77 & \underline{82.74} & 81.91 & \textbf{83.88} & \textcolor{blue}{+1.97} \\
\bottomrule
\end{tabular}%
}
\vspace{-0.5em}
\caption{\textnormal{Base-to-novel generalization results on 11 datasets. `Base' and `Novel' indicate accuracy on base and novel classes, respectively, while `HM' denotes their harmonic mean. `Gain' shows the improvement of GLAD over the CLIP-LoRA baseline. Our method achieves clear gains over the baseline and demonstrates strong generalization ability, achieving the best or second best HM on all datasets. \textbf{Bold} indicates the best result, \underline{underline} indicates the second best. Blue indicates improvement and red indicates decrease.}}
\label{tab:base2novel}
\vspace{-1.5em}
\end{table*}

\vspace{-0.2em}
\subsection{Experimental Settings}
\vspace{-0.2em}

\begin{table*}[t]
    % \tabstyle{2.5pt}
    \scalebox{0.95}{
    \begin{tabular}{l c ccccccccccc}
    \toprule
    & \textbf{Source} & \multicolumn{11}{c}{\textbf{Target}} \\ \cmidrule(lr){2-2} \cmidrule(lr){3-13}
    & \rotatebox{90}{ImageNet} & \rotatebox{90}{Caltech101} & \rotatebox{90}{OxfordPets} & \rotatebox{90}{StanfordCars} & \rotatebox{90}{Flowers102} & \rotatebox{90}{Food101} & \rotatebox{90}{Aircraft} & \rotatebox{90}{SUN397} & \rotatebox{90}{DTD} & \rotatebox{90}{EuroSAT} & \rotatebox{90}{UCF101} & \rotatebox{90}{\emph{Average}} \\
    \midrule
    CoOp & 71.51 & 93.70 & 89.14 & 64.51 & 68.71 & 85.30 & 18.47 & 64.15 & 41.92 & 46.39 & 66.55 & 63.88 \\
    
    Co-CoOp & 71.02 &\textbf{94.43} & 90.14 & 65.32 & \underline{71.88} & 86.06 & 22.94 & \underline{67.36} & 45.73 & 45.37 & 68.21 & 65.74 \\
    
    MaPLe & 70.72 & 93.53 & \underline{90.49} & 65.5 & \textbf{72.23} & \underline{86.20} & \textbf{24.74} & 67.01 & 46.49 & \textbf{48.06} & 68.69 & \underline{66.30} \\
    
    PromptSRC & 71.27 & 93.60 & 90.25 & \underline{65.70} & 70.25 & 86.15 & \underline{23.90} & 67.10 & \underline{46.87} & 45.50 & \underline{68.75} & 65.81 \\

    CLIP-LoRA & \textbf{73.00} & 92.50 & 88.13 & 57.77 & 64.67 & 81.90 & 20.03 & 64.30 & 42.07 & 44.30 & 64.50 & 62.02 \\

    \midrule

    GLAD (Ours) & \underline{71.83} & \underline{93.97} & \textbf{90.83} & \textbf{66.43} & \textbf{72.23} & \textbf{86.47} & 23.83 & \textbf{68.60} & \textbf{47.10} & \underline{47.73} & \textbf{69.27} & \textbf{66.65} \\
    Gain ($\Delta$) & \textcolor{red}{-1.17} & \textcolor{blue}{+1.47} & \textcolor{blue}{+2.70} & \textcolor{blue}{+8.66} & \textcolor{blue}{+7.56} & \textcolor{blue}{+4.57} & \textcolor{blue}{+3.80} & \textcolor{blue}{+4.30} & \textcolor{blue}{+5.03} & \textcolor{blue}{+3.43} & \textcolor{blue}{+4.77} & \textcolor{blue}{+4.63} \\
    \bottomrule
    \end{tabular}}
    \vspace{-0.5em}
    \caption{\textnormal{Cross-dataset generalization results. All models are trained on ImageNet and directly evaluated on 11 unseen datasets without further adaptation. `Gain' shows the improvement of GLAD over the LoRA baseline. Our method significantly improves cross-dataset generalization over the baseline, achieving the best or second best result on 9 out of 10 target datasets, and obtains the highest overall average accuracy across all target domains. \textbf{Bold} indicates the best result, \underline{underline} indicates the second best result, blue indicates improvement, and red indicates decrease compared to the baseline CLIP-LoRA.}}
\label{tab:xd}
\vspace{-1.5em}
\end{table*}

\textbf{Datasets.} We conduct experiments on eleven widely-used image classification benchmarks spanning a diverse range of domains and difficulty levels. These include: ImageNet \cite{deng2009imagenet}, Caltech101 \cite{fei2004learning}, Oxford-Pets \cite{parkhi2012cats}, Stanford-Cars \cite{krause20133d}, Flowers102 \cite{nilsback2008automated}, Food101 \cite{bossard2014food}, FGVC-Aircraft \cite{maji2013fine}, SUN397 \cite{xiao2010sun}, UCF101 \cite{soomro2012ucf101}, DTD \cite{cimpoi2014describing}, and EuroSAT \cite{helber2019eurosat}. These datasets cover a wide spectrum, including object recognition, fine-grained classification, scene understanding, texture analysis, and satellite imagery.

\textbf{Evaluation Protocol.}
To comprehensively evaluate model generalization, we conduct experiments under three settings following \cite{huang2022learning, zhou2022conditional}: (1) base-to-novel class generalization, where the data is split into base and novel subsets by category, and models are trained on base classes with 16-shot supervision and tested on unseen novel classes; (2) image domain generalization, where models trained on ImageNet are evaluated on four domain-shifted variants; and (3) cross-dataset evaluation, where ImageNet-trained models are tested on ten diverse classification benchmarks without additional fine-tuning.

\textbf{Baselines.} We compare GLAD with zero-shot CLIP \cite{radford2021learning} and a series of strong prompt tuning methods, including CoOp \cite{zhou2022learning}, CoCoOp \cite{zhou2022conditional}, ProDA \cite{lu2022prompt}, RPO \cite{lee2023read}, MaPLe~\cite{khattak2023maple}, PromptSRC \cite{khattak2023self}, as well as a baseline CLIP-LoRA \cite{zanella2024low} that is trained using the original LoRA on CLIP. All of these baselines are trained under the same few-shot splits and training configurations for fair comparison.

\textbf{Implementation Details.} We implement LoRA with rank $r=8$ for all inserted modules. AlignNet is designed as an MLP with hidden dimensions 256 and 128. For gradient regularization, we set regularization strength $\alpha=0.5$ and perturbation strength $\rho=0.1$. Training runs for 20 epochs with a cosine annealing learning rate schedule, starting from 0.001. The CLIP backbone is frozen to retain pretrained knowledge. All experiments use a single NVIDIA RTX 3090 GPU (24GB memory). Results are averaged over three random seeds for robustness and reproducibility.

% \begin{table}[h]
%     \centering
%     \small
%     \resizebox{\columnwidth}{!}{
%     \begin{tabular}{l cccccc}
%         \toprule
%         & \textbf{Source} & \multicolumn{5}{c}{\textbf{Target}} \\
%         \cmidrule(lr){2-2} \cmidrule(lr){3-7}
%         & ImageNet & -V2 & -S & -A & -R & Avg. \\
%         \midrule
%         CLIP & 66.73 & 60.83 & 46.15 & 47.77 & 73.96 & 57.18 \\
%         CoOp & 71.51 & 64.20 & 47.99 & 49.71 & 75.21 & 59.28 \\
%         Co-CoOp & 71.02 & 64.07 & 48.75 & 50.63 & 76.18 & 59.91 \\
%         MaPLe & 70.72 & 64.07 & 49.15 & \underline{50.90} & 76.98 & 60.27 \\
%         PromptSRC & 71.27 & 64.35 & \underline{49.55} & \underline{50.90} & \textbf{77.80} & \underline{60.65} \\
        
%         \midrule
        
%         LoRA (Baseline) & \textbf{73.00} & \textbf{65.17} & 45.97 & 44.17 & 72.67 & 57 \\
%         GLAD (Ours) & \underline{71.83} & \underline{64.93} & \textbf{49.8} & \textbf{51.03} & \underline{77.43} & \textbf{60.8} \\
%         \bottomrule
%     \end{tabular}}
%     \vspace{0.5em}
%     \caption{\textnormal{Domain generalization results. All methods are trained on ImageNet and evaluated on four out-of-distribution variants: ImageNet-V2, -Sketch, -A, and -R. GLAD achieves a substantial lead on the source domain (ImageNet) and obtains the highest average accuracy across target domains, demonstrating strong robustness to distribution shifts.}}
%     \label{tab:dg}
%     \vspace{-1.5em}
% \end{table}

\begin{table}[h]
    \centering
    \small
    \resizebox{\columnwidth}{!}{
    \begin{tabular}{l cccccc}
        \toprule
        & \textbf{Source} & \multicolumn{5}{c}{\textbf{Target}} \\
        \cmidrule(lr){2-2} \cmidrule(lr){3-7}
        & ImageNet & -V2 & -S & -A & -R & Avg. \\
        \midrule
        CLIP & 66.73 & 60.83 & 46.15 & 47.77 & 73.96 & 57.18 \\
        CoOp & 71.51 & 64.20 & 47.99 & 49.71 & 75.21 & 59.28 \\
        Co-CoOp & 71.02 & 64.07 & 48.75 & 50.63 & 76.18 & 59.91 \\
        MaPLe & 70.72 & 64.07 & 49.15 & \underline{50.90} & 76.98 & 60.27 \\
        PromptSRC & 71.27 & 64.35 & \underline{49.55} & \underline{50.90} & \textbf{77.80} & \underline{60.65} \\
        CLIP-LoRA & \textbf{73.00} & \textbf{65.17} & 45.97 & 44.17 & 72.67 & 57.00 \\
        
        \midrule
        GLAD (Ours) & \underline{71.83} & \underline{64.93} & \textbf{49.80} & \textbf{51.03} & \underline{77.43} & \textbf{60.80} \\
        Gain ($\Delta$) & \textcolor{red}{-1.17} & \textcolor{red}{-0.24} & \textcolor{blue}{+3.83} & \textcolor{blue}{+6.86} & \textcolor{blue}{+4.76} & \textcolor{blue}{+3.80} \\
        \bottomrule
    \end{tabular}}
    \vspace{-0.5em}
    \caption{\textnormal{Domain generalization results. All methods are trained on ImageNet and evaluated on four out-of-distribution variants: ImageNet-V2, -Sketch, -A, and -R. `Gain' shows the improvement of GLAD over the CLIP-LoRA baseline. GLAD achieves a substantial lead in target domain average accuracy and shows strong robustness to distribution shifts. \textbf{Bold} indicates the best result, \underline{underline} indicates the second best, blue indicates improvement and red indicates decrease compared to the baseline.}}
    \label{tab:dg}
    \vspace{-1.5em}
\end{table}

\vspace{-0.2em}
\subsection{Main Results}
\vspace{-0.2em}
\textbf{Base-to-Novel Generalization.}
We evaluate GLAD on the base-to-novel class generalization task, where each dataset is split into disjoint base and novel categories, and models are trained with 16-shot supervision on base classes. We compare against zero-shot CLIP~\cite{radford2021learning}, CLIP-LoRA baseline, and several prompt-based methods, including CoOp~\cite{zhou2022learning}, CoCoOp~\cite{zhou2022conditional}, ProDA~\cite{lu2022prompt}, RPO~\cite{lee2023read}, MaPLe~\cite{khattak2023maple}, and PromptSRC~\cite{khattak2023self}. Results on 11 benchmarks are shown in Table~\ref{tab:base2novel}.

GLAD achieves the highest average accuracy on both base (85.05\%) and novel (76.74\%) classes, significantly improving over the LoRA baseline (+1.67\% HM gain) and surpassing prior methods. It attains the best average HM of 80.68\%, outperforming the strongest baseline PromptSRC by 0.71\% in HM, demonstrating balanced generalization across seen and unseen classes.

Across datasets, GLAD ranks first in HM on 7 out of 11 datasets and second on the rest, showing strong performance not only on large benchmarks like ImageNet and Caltech101, but also on challenging domains such as FGVC Aircraft, DTD, and SUN397. These results highlight GLAD’s clear advantage in base-to-novel generalization and its state-of-the-art accuracy across diverse tasks.

\textbf{Domain Generalization.}
To evaluate GLAD’s robustness under distribution shifts, we follow standard domain generalization protocol: all models are trained on ImageNet~\cite{deng2009imagenet} and directly tested on four out-of-distribution (OOD) variants: ImageNet-V2~\cite{recht2019imagenet}, ImageNet-Sketch~\cite{wang2019learning}, ImageNet-A~\cite{hendrycks2021natural}, and ImageNet-R~\cite{hendrycks2021many} without additional fine-tuning. These target datasets introduce diverse domain shifts, such as low-level distortions (Sketch), adversarial filtering (A), and artistic style variations (R).

As shown in Table~\ref{tab:dg}, GLAD achieves clear improvements over the CLIP-LoRA baseline on target domains, boosting average target accuracy by +3.80\%. Notably, GLAD achieves the best accuracy on ImageNet-Sketch (49.80\%), ImageNet-A (51.03\%), and the highest overall target average (60.80\%), outperforming prior prompt-based methods including MaPLe (60.27\%) and PromptSRC (60.65\%). It also secures competitive performance on ImageNet-R (77.43\%) and ImageNet-V2 (64.93\%).

These results demonstrate that GLAD not only enhances performance over CLIP-LoRA \cite{zanella2024low} under domain shifts but also provides robust and generalizable vision-language adaptation compared to existing prompt tuning methods.

\textbf{Cross-Dataset Generalization.}
We evaluate GLAD’s transferability by training on ImageNet~\cite{deng2009imagenet} and directly testing on ten out-of-distribution (OOD) datasets without further adaptation. These datasets cover diverse domains, including objects, scenes, textures, fine-grained categories, and actions, offering a comprehensive test of cross-dataset generalization.

As shown in Table~\ref{tab:xd}, GLAD achieves substantial improvements over the CLIP-LoRA baseline, boosting average accuracy across target domains by +4.63\%. It achieves the best or second-best accuracy on 9 out of 10 target datasets, demonstrating robust generalization across various domains. Notably, GLAD delivers large gains on challenging datasets like StanfordCars (+8.66\%), Flowers102 (+7.56\%), and DTD (+5.03\%). Compared to existing prompt-based methods, GLAD consistently achieves higher accuracy, including the highest average target accuracy (66.65\%), surpassing MaPLe (66.30\%) and PromptSRC (65.81\%).

These results highlight that GLAD provides significant advantages over LoRA and prior prompt tuning methods, offering a strong and reliable solution for cross-domain transfer in vision-language models.

\vspace{-0.4em}
\subsection{Ablation Studies}
\vspace{-0.4em}
To better understand the contribution of each GLAD component, we conduct ablation studies, with results shown in Table~\ref{tab:ablations}. We analyze the individual and combined effects of SAM, Gradient Regularization, and AlignNet, and provide detailed insights into their impact.

\begin{table}[!]
     \centering
 \setlength{\tabcolsep}{3pt}
    \resizebox{0.98\columnwidth}{!}{
    \begin{tabular}{c|c c c c| ccc }
    \toprule
    & LoRA & SAM & GradReg & AlignNet & Base & Novel & H\\
    \midrule
     (a) & \cmark & & & & 84.47 & 74.22 & 79.01 \\
     (b) & \cmark & \cmark & & & 83.79 & 76.54 & 80.00 \\
     (c) & \cmark &  & \cmark & & 84.60 & 76.57 & 80.38 \\
     (d) &\cmark & & & \cmark & 84.82 & 74.76 & 79.81 \\
     (e) & \cmark &  & \cmark & \cmark & 85.05 & 76.74 & 80.68 \\
    \bottomrule
    \end{tabular}
    }
    \vspace{-0.5em}
    \caption{\textnormal{Ablation study on individual GLAD components. We report average performance over 11 datasets. {LoRA} denotes low-rank adaptation for internal representation tuning and serves as the baseline model. {GradReg} refers to the gradient regularization strategy. {AlignNet} adjusts static text embeddings conditioned on dynamic visual features. Compared to the LoRA baseline, SAM improves novel accuracy but slightly reduces base accuracy. Gradient Regularization retains SAM’s benefits for novel classes while preserving base performance. AlignNet enhances both base and novel accuracy, and combining AlignNet with GradReg achieves the best overall performance. `Base', `Novel', and `H' refer to accuracy on base classes, novel classes, and their harmonic mean, respectively.}}
    \label{tab:ablations}
    \vspace{-1.5em}
\end{table}

\textbf{Effectiveness of SAM.}  
As shown in row (b), adding SAM to the LoRA baseline improves novel accuracy from 74.22\% to 76.54\%, confirming that SAM helps the model find flatter minima that generalize better to unseen categories. However, this comes at the cost of a slight drop in base accuracy (from 84.47\% to 83.79\%). This suggests that although SAM enhances generalization, its optimization landscape may limit the model’s ability to fully exploit the training data in low-shot settings, making convergence to low-loss regions on base classes more difficult.

\textbf{Effectiveness of Gradient Regularization.}  
In row (c), replacing SAM with Gradient Regularization raises novel accuracy further to 76.57\% while recovering base accuracy to 84.60\%. This shows that Gradient Regularization inherits SAM’s generalization advantage without sacrificing learning capacity. The strategy promotes smoother solutions that generalize well, while still allowing the model to effectively fit the training data on base classes.

\textbf{Effectiveness of AlignNet.}  
Row (d) demonstrates that adding AlignNet to LoRA improves both base and novel accuracy (84.82\% and 74.76\%, respectively). AlignNet introduces dynamic adjustments to static text embeddings based on visual inputs, providing additional flexibility that allows the model to better align vision and language features. This helps the model adapt more effectively to both seen and unseen categories, leading to balanced performance.

\textbf{Synergy of All Components.}  
Finally, the full GLAD model (row e), which combines LoRA, Gradient Regularization, and AlignNet, achieves the best overall performance: 85.05\% base accuracy, 76.74\% novel accuracy, and 80.68\% harmonic mean. This indicates that the components complement each other: Gradient Regularization ensures generalization without harming learning ability, while AlignNet further enhances flexibility and alignment between modalities. Their combination enables GLAD to achieve robust and balanced performance across base and novel classes in few-shot scenarios.

These findings highlight that each component plays a distinct role: SAM and Gradient Regularization primarily improve generalization, AlignNet provides adaptability, and together they create a stronger model that balances learning and generalization under challenging conditions.

\vspace{-0.5em}

% \begin{figure}[!]
%     \centering
%     \includegraphics[width=0.9\columnwidth]{Figures/lora_rank.pdf}
%     \vspace{-3.5pt}
%     \caption{\textnormal{This figure illustrates how different LoRA ranks affect the base-to-novel harmonic mean (H) when LoRA is used alone.}}
%     \label{fig:lora_rank}
% \end{figure}

% \textbf{Effect of LoRA Rank.} To further analyze the impact of LoRA rank, we visualize the base-to-novel harmonic mean (H) achieved under different rank values in Figure~\ref{fig:lora_rank}. The results show that the performance improves significantly as the rank increases from 2 to 8, reaching a peak H of 79.08\% at rank=8. However, further increasing the rank to 16 leads to a slight drop in performance. This suggests that while increasing LoRA rank provides more capacity for adaptation, excessive capacity may introduce overfitting under few-shot settings. Therefore, rank=8 offers the best trade-off between expressiveness and generalization in our setting.

% \section{Conclusion and Discussion}
\vspace{-0.4em}
\section{Conclusion}
\vspace{-0.4em}

% We propose GLAD, a generalizable and parameter-efficient fine-tuning framework for vision-language models. GLAD integrates low-rank adaptation, lightweight multi-modal fusion, and gradient-based regularization to improve both task-specific adaptation and generalization. By freezing the CLIP backbone and updating only a small set of auxiliary parameters, GLAD achieves strong robustness to distribution shifts and consistently outperforms existing prompt tuning methods under few-shot settings. Ablation studies confirm that LoRA, MetaNet, and gradient regularization each contribute effectively and work synergistically. Overall, GLAD offers a simple yet effective solution for adapting large-scale vision-language models under limited supervision while preserving their generalization ability.

We present GLAD, a parameter-efficient fine-tuning framework designed to enhance the generalization of vision-language models. GLAD integrates three key components: a low-rank adaptation baseline, a lightweight multi-modal fusion module, and a gradient-based regularization strategy. By updating only a small set of parameters while keeping the CLIP backbone frozen, GLAD achieves robust performance under distribution shifts and consistently outperforms existing prompt tuning approaches in few-shot scenarios. Our ablation analysis examines each component’s contribution and shows how Gradient Regularization overcomes the limitations of SAM in low-shot settings. Overall, GLAD offers an effective, efficient solution for adapting vision-language models with limited supervision while ensuring strong generalization across tasks.
\vspace{-0.5em}

% While GLAD demonstrates strong performance in both adaptation and generalization benchmarks, there remains room for further improvement. First, future work may explore more expressive low-rank modules, such as adaptive or task-aware variants of LoRA, to enhance the flexibility of fine-tuning. Additionally, more advanced fusion strategies—such as cross-modal attention or deeper interaction modules—could further strengthen semantic alignment between modalities. Finally, although our gradient regularization method is task-agnostic by design, we validate its effectiveness only in the context of CLIP fine-tuning. Future studies may apply this regularization strategy to other domains to verify its broader applicability and performance.

{
    \small
    \bibliographystyle{ieeenat_fullname}
    \bibliography{main}
}

\end{document}